\documentclass[runningheads]{llncs}
\usepackage[T1]{fontenc}
\usepackage{graphicx}
\usepackage{bm}
\usepackage{booktabs}
\usepackage{times}
\usepackage{multirow}
\usepackage{lscape}
\usepackage[ruled,linesnumbered]{algorithm2e}
\usepackage{subfigure}
\usepackage{enumitem}
\usepackage{hyperref}
\usepackage{dirtytalk}
\usepackage[misc]{ifsym}
\begin{document}
\title{Continually learning out-of-distribution spatiotemporal data for robust energy forecasting}
\titlerunning{CL, OoD, and ST data for robust energy forecasting}
\toctitle{Continually learning out-of-distribution spatiotemporal data for robust energy forecasting}

\author{
Arian Prabowo\inst{1}\Letter\orcidID{0000-0002-0459-354X} \and
Kaixuan Chen\inst{1} \and
Hao Xue \inst{1}\orcidID{0000-0003-1700-9215} \and
Subbu Sethuvenkatraman \inst{2}\orcidID{0000-0001-7197-2307} \and
Flora D. Salim \inst{1}\orcidID{0000-0002-1237-1664}} 
\authorrunning{A. Prabowo et al.}
%
\institute{
UNSW, Sydney, Australia\\
\email{\{arian.prabowo,
         kaixuan.chen,
         hao.xue1
         flora.salim\}@UNSW.edu.au} \and
CSIRO, Newcastle, Australia
\email{subbu.sethuvenkatraman@csiro.au}
}

\tocauthor{
	Arian~Prabowo,
	Kaixuan~Chen,
	Hao~Xue,
	Subbu~Sethuvenkatraman, and
	Flora~D.~Salim
}

\maketitle              
\begin{abstract}
Forecasting building energy usage is essential for promoting sustainability and reducing waste, as it enables building managers to adjust energy use to improve energy efficiency and reduce costs. This importance is magnified during anomalous periods, such as the COVID-19 pandemic, which have disrupted occupancy patterns and made accurate forecasting more challenging.
Forecasting energy usage during anomalous periods is difficult due to changes in occupancy patterns and energy usage behavior. One of the primary reasons for this is the shift in distribution of occupancy patterns, with many people working or learning from home. This has created a need for new forecasting methods that can adapt to changing occupancy patterns.
Online learning has emerged as a promising solution to this challenge, as it enables building managers to adapt to changes in occupancy patterns and adjust energy usage accordingly. With online learning, models can be updated incrementally with each new data point, allowing them to learn and adapt in real-time.
Continual learning methods offer a powerful solution to address the challenge of catastrophic forgetting in online learning, allowing energy forecasting models to retain valuable insights while accommodating new data and improving generalization in out-of-distribution scenarios.
Another solution is to use human mobility data as a proxy for occupancy, leveraging the prevalence of mobile devices to track movement patterns and infer occupancy levels. Human mobility data can be useful in this context as it provides a way to monitor occupancy patterns without relying on traditional sensors or manual data collection methods.
We have conducted extensive experiments using data from four buildings to test the efficacy of these approaches. However, deploying these methods in the real world presents several challenges. 

\end{abstract}

\section{Introduction}
Accurate prediction of the electricity demand of buildings is vital for effective and cost-efficient energy management in commercial buildings. It also plays a significant role in maintaining a balance between electricity supply and demand in modern power grids. However, forecasting energy usage during anomalous periods, such as the COVID-19 pandemic, can be challenging due to changes in occupancy patterns and energy usage behavior. One of the primary reasons for this is the shift in distribution of occupancy patterns, with many people working or learning from home, leading to increased residential occupancy and decreased occupancy in offices, schools, and most retail establishments.
Essential retail stores, such as grocery stores and restaurants, might experience a divergence between occupancy and energy usage, as they have fewer dine-in customers but still require energy for food preparation and sales.
This has created a need for new forecasting methods that can adapt to changing occupancy patterns.

Online learning has emerged as a promising solution to this challenge, as it enables building managers to adapt to changes in occupancy patterns and adjust energy usage accordingly. With Online learning, models can be updated incrementally with each new data point, allowing them to learn and adapt in real-time \cite{hoi2021OLsurvey}.

Furthermore, continual learning methods offer an even more powerful solution by addressing the issue of catastrophic forgetting~\cite{lin1992ER,chaudhry2019ER}.
These methods allow models to retain previously learned information while accommodating new data, preventing the loss of valuable insights and improving generalization in out-of-distribution scenarios. By combining online learning with continual learning techniques, energy forecasting models can achieve robustness, adaptability, and accuracy, making them well-suited for handling the challenges posed by spatiotemporal data with evolving distributions.

Another solution is to use human mobility data as a proxy for occupancy, leveraging the prevalence of mobile devices to track movement patterns and infer occupancy levels. Human mobility data can be useful in this context as it provides a way to monitor occupancy patterns without relying on traditional sensors or manual data collection methods.~\cite{Salim2020-fc}

In this study, we evaluate the effectiveness of mobility data and continual learning for forecasting building energy usage during anomalous periods. We utilized real-world data from Melbourne, Australia, a city that experienced one of the strictest lockdowns globally~\cite{boaz_2021}, making it an ideal case for studying energy usage patterns during out-of-distribution periods. We conducted experiments using data from four building complexes to empirically assess the performance of these methods.

\section{Related Works}






\subsection{Energy Prediction in Urban Environments}
Electricity demand profiling and forecasting has been a task of importance for many decades. Nevertheless, there exist a limited number of work in literature that investigate how human mobility patterns are directly related to the urban scale energy consumption, both during normal periods as well as adverse/extreme events. Energy modelling in literature is done at different granularities, occupant-level (personal energy footprinting), building-level and city-level. Models used for energy consumption prediction in urban environments are known as Urban Building Energy Models (UBEM). While top-down UBEMs are used for predicting aggregated energy consumption in urban areas using macro-economic variables and other aggregated statistical data, bottom-up UBEMs are more suited for building-level modelling of energy by clustering buildings into groups of similar characteristics~\cite{ALI2021111073}. Some examples in this respect are SUNtool, CitySim, UMI, CityBES, TEASER and HUES. Software modelling (simulation-based) is also a heavily used approach for building-wise energy prediction (Eg: EnergyPlus~\cite{CRAWLEY2001319}). Due to fine-grain end-user level modelling, bottom-up UBEMs can incorporate inputs of occupant schedules. There also exist occupant-wise personal energy footprinting systems. However, for such occupant-wise energy footprinting, it requires infrastructure related to monitoring systems and sensors for indoor occupant behaviours, which are not always available. Also, due to privacy issues, to perform modelling at end-user level granularity, it can be hard to get access to publicly available data at finer temporal resolutions (both occupancy and energy)~\cite{Wei2019-hw}. Building-wise energy models also have the same problems. Simulation-based models have complexity issues when scaling to the city level, because they have to build one model per each building. Moreover, simulation-based models contain assumptions about the data which make their outputs less accurate~\cite{Ali2020-me}. Consequently, it remains mostly an open research area how to conduct energy forecasting with data distribution shifts.

\subsection{Mobility Data as Auxiliary Information in Forecasting}
The study of human mobility patterns involves analysing the behaviours and movements of occupants in a particular area in a spatio-temporal context~\cite{Salim2020-fc}. The amount of information that mobility data encompasses can be huge. The behaviour patterns of humans drive the decision making in many use-cases. Mobility data in particular, can act as a proxy for the dynamic (time varying) human occupancy at various spatial densities (building-wise, city-wise etc.). Thus such data are leveraged extensively for many tasks in urban environments including predicting water demand~\cite{Smolak2020-xc}, urban flow forecasting~\cite{Xue2021-br}, predicting patterns in hospital patient rooms~\cite{DEDESKO2015136}, electricity use~\cite{hansika2023human} etc.\ that depend on human activities.

Especially, during the COVID19 pandemic, mobility data has been quite useful for disease propagation modelling. For example, in the work by \cite{Wang2020-hi}, those authors have developed a Graph Neural Network (GNN) based deep learning architecture to forecast the daily new COVID19 cases state-wise in United States. The GNN is developed such that each node represents one region and each edge represents the interaction between the two regions in terms of mobility flow. The daily new case counts, death counts and intra-region mobility flow is used as the features of each node whereas the inter-region mobility flow and flow of active cases is used as the edge features. Comparisons against other classical models which do not use mobility data has demonstrated the competitiveness of the developed model.

Nevertheless, as \cite{Salim2020-fc} state, the existing studies involving human mobility data lack diversity in the datasets in terms of their social demographics, building types, locations etc. Due to the heterogeneity, sparsity and difficulty in obtaining diverse mobility data, it remains a significant research challenge to incorporate them in modelling techniques~\cite{ALI2021111073}. Yet, the lack of extracting valuable information from such real-world data sources remains untapped, with a huge potential of building smarter automated decision making systems for urban planning~\cite{Salim2020-fc}.

\subsection{Deep Learning for Forecasting}

Deep learning has gained significant popularity in the field of forecasting, with various studies demonstrating its effectiveness in different domains~\cite{Herzen2022Darts}.
For instance, it has been widely applied in mobility data forecasting, including road traffic forecasting~\cite{prabowo2023GSWaN,prabowo2023MPNN4TrafficForecasting,PRABOWOArian2022StDL}, and flight delay forecasting~\cite{shao2022predicting}.
In the realm of electricity forecasting, Long Short-Term Memory (LSTM) networks have been widely utilized~\cite{pelka2020lstmLoadForecasting}.
Another popular deep learning model for electricity load forecasting is Neural basis expansion analysis for interpretable time series forecasting (N-BEATS)~\cite{ORESHKIN2021116918NBEATSenergyForecasting}.

However, one common challenge faced by these deep learning methods is the performance degradation when the data distributions change rapidly, especially during out-of-distribution (OOD) periods.
Online learning methods have been proposed to address this issue~\cite{mahadik2020fast,li2016art,kar2016online}.
However, online learning methods can suffer from catastrophic forgetting, where newly acquired knowledge erases previously learned information~\cite{Salim2020-fc}.
To mitigate this, continual learning methods have been developed, which aim to retain previously learned information while accommodating new data, thereby improving generalization in OOD scenarios.

One approach to continual learning is Experience Replay~\cite{lin1992ER,chaudhry2019ER}, a technique that re-exposes the model to past experiences to improve learning efficiency and reduce the effects of catastrophic forgetting.
Building upon this idea, the Dark Experience Replay++ algorithm~\cite{buzzega2020derpp} utilizes a memory buffer to store past experiences and a deep neural network to learn from them, employing a dual-memory architecture that allows for the storage of both short-term and long-term memories separately.
Another approach is the Fast and Slow Network (FSNet)~\cite{pham2022learning}, which incorporates a future adaptor and an associative memory module. The future adaptor facilitates quick adaptation to changes in the data distribution, while the associative memory module retains past patterns to prevent catastrophic forgetting.
These continual learning methods have shown promise in mitigating catastrophic forgetting and improving generalization in OOD scenarios.

In the context of energy forecasting, the utilization of continual learning techniques holds great potential for addressing the challenges posed by OOD spatiotemporal data.
By preserving past knowledge and adapting to new patterns, these methods enable more robust and accurate energy forecasting even during periods of rapid data distribution shifts.

\section{Problem Definition}
\subsection{Time Series Forecasting}

Consider a multivariate time series $\mathcal{X}\in \mathbf{R}^{T\times N}$ comprising mobility data, weather data, and the target variable, which is the energy consumption data. The time series consists of $T$ observations and $N$ dimensions. To perform $H$-timestamps-ahead time series forecasting, a model $f$ takes as input a look-back window of $L$ historical observations $(\mathbf{x}_{t-L+1},\mathbf{x}_{t-L+2},...,\mathbf{x}_{t})$ and generates forecasts for $H$ future observations of the target variable $y$, which corresponds to the energy consumption of a building.
We have:
\begin{equation}
    f_{\bm{\omega}}(\mathbf{x}_{t-L+1},\mathbf{x}_{t-L+2},...,\mathbf{x}_{t}) =
   (y_{t+1},y_{t+2},...,y_{t+H}),
\end{equation}
where $\omega$ denotes the parameters in the model.

\subsection{Continual Learning for Time Series Forecasting}
In a continual learning setting, the conventional machine learning practice of separating data into training and testing sets with a $70\%$ to $30\%$ ratio does not apply, as learning occurs continuously over the entire period. After an initial pre-training phase using a short period of training data, typically the first 3 months, the model continually trains on incoming data and generates predictions for future time windows. Evaluation of the model's performance is commonly done by measuring its accumulated errors throughout the entire learning process \cite{sahoo2018online}. 

\section{Method}

\begin{figure}[t]
\centering
\includegraphics[width=\textwidth]{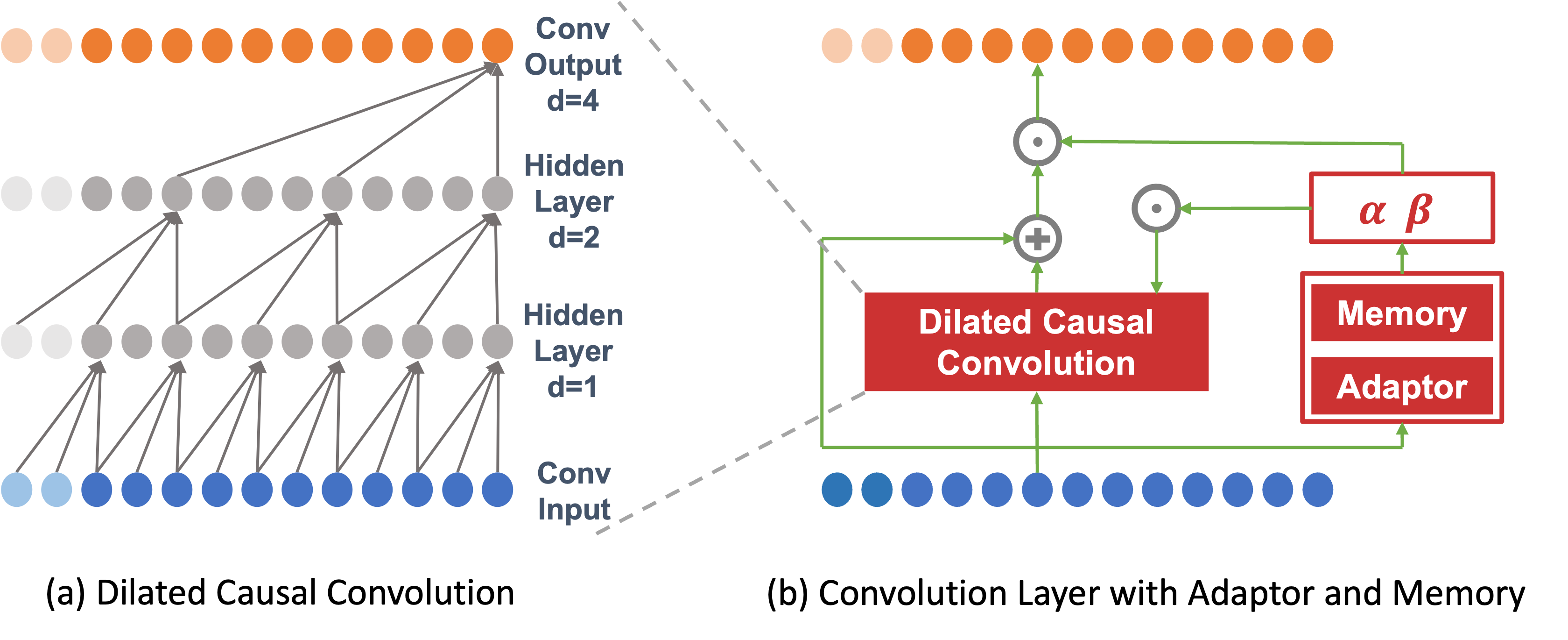}
\caption{The convolution architecture in TCN}
\label{fig:architecture}
\end{figure}

Continual learning presents unique challenges that necessitate the development of specialized algorithms and evaluation metrics to address the problem effectively. In this context, a continual learner must strike a balance between retaining previously acquired knowledge while facilitating the learning of new tasks. In time-series forecasting, the challenge lies in balancing the need to learn new temporal dependencies quickly while remembering past patterns, a phenomenon commonly referred to as the stability-plasticity dilemma \cite{grossberg2013adaptive}. Building on the concept of complementary learning systems theory for dual learning systems \cite{kumaran2016learning}, a Temporal Convolutional Network (TCN) is utilized as the underlying architecture, which is pre-trained to extract temporal features from the training dataset. Subsequently, the convolutional layers of the TCN are customized with a future adaptor and an associative memory module to address the challenges associated with continual learning. The future adaptor facilitates quick adaptation to changes, while the associative memory module is responsible for retaining past patterns to prevent catastrophic forgetting.
In this section we describe in detail the architecture of FSNet~\cite{pham2022learning}.

\subsection{Backbone-Temporal Convolutional Network}

FSNet adopts the TCN proposed by Bai et al. \cite{bai2018empirical} as the backbone architecture for extracting features from time series data. Although traditional Convolutional Neural Networks (CNNs) have shown great success in image-processing tasks, their performance in time-series forecasting is often unsatisfactory. This is due to several reasons, including (a) the difficulty of capturing contextual relationships using CNNs, (b) the risk of information leakage caused by traditional convolutions that incorporate future temporal information, and (c) the loss of detail associated with pooling layers that extract contour features. In contrast, TCN's superiority over CNNs can be attributed to its use of causal and dilated convolutions, which enhance its ability to capture temporal dependencies in a more effective manner.

\subsubsection{Causal Convolutions}

In contrast to traditional CNNs, which may incorporate future temporal information and violate causality, causal convolutions are effective in avoiding data leakage in the future. By only considering information up to and including the current time step, causal convolutions do not alter the order in which data is modelled and are therefore well-suited for temporal data. 
Specifically, to ensure that the output tensor has the same length as the input tensor, it is necessary to perform zero-padding. When zero-padding is performed only on the left side of the input tensor, causal convolution can be ensured. In Figure~\ref{fig:architecture} (a), zero-padding is shown in light colours on the left side. There is no padding on the right side of the input sequence because the last element of the input sequence is the latest element on which the rightmost output element depends. Regarding the second-to-last output element, its kernel window is shifted one position to the left compared to the last output element. This implies that the second-to-last element's latest dependency on the rightmost side of the input sequence is the second-to-last element. By induction, for each element in the output sequence, its latest dependency in the input sequence has the same index as the element itself.

\subsubsection{Dilated Convolutions.}

Dilated convolution is an important component of TCN because causal convolution can only access the past inputs up to a certain depth, which is determined by the kernel size of the convolutional layer. In a deep network, the receptive field of the last layer may not be large enough to capture long-term dependencies in the input sequence. In dilated convolutions, the dilation factor is used to determine the spacing between the values in the kernel of the dilated convolution.
More formally, we have: 

\begin{equation}
Conv(\mathbf{x})_{i} = \sum_{m=0}^{k}w_m\cdot\mathbf{x}_{i-m\times d}
\end{equation}
where $i$ represents the $i$-th element, $w$ denotes the kernel, $d$ is the dilation factor, $k$ is the filter size.
Dilation introduces a fixed step between adjacent filter taps. Specifically, if the dilation factor $d$ is set to 1, the dilated convolution reduces to a regular convolution. However, for $d > 1$, the filters are expanded by $d$ units, allowing the network to capture longer-term dependencies in the input sequence. A dilated causal convolution architecture can be seen in Figure~\ref{fig:architecture} (a).

\subsection{Fast Adaptation}

FSNet modify the convolution layer in TCN to achieve fast adaptation and associative memory. The modified structure is illustrated in Figure~\ref{fig:architecture} (b). In this subsection, we first introduce the fast adaptation module.

In order to enable rapid adaptation to changes in data streams and effective learning with limited data, Sahoo et al. \cite{sahoo2018online} and Phuong and Lampert \cite{phuong2019distillation} propose the use of shallower networks and single layers that can quickly adapt to changes in data streams or learn more efficiently with limited data. Instead of limiting the depth of the network, it is more advantageous to enable each layer to adapt independently. In this research, we adopt an independent monitoring and modification approach for each layer to enhance the learning of the current loss. An adaptor is utilized to map the recent gradients of the layer to a smaller, more condensed set of transformation parameters to adapt the backbone. However, the gradient of a single sample can cause significant fluctuation and introduce noise into the adaptation coefficients in continual time-series forecasting. As a solution, we utilize Exponential Moving Average (EMA) gradient to mitigate the noise in online training and capture the temporal information in time series:
\begin{equation}
\hat{g_l} = \gamma \hat{g_l} + (1 - \gamma)\hat{g_l^t},
\label{eq:ema}
\end{equation}
where $\hat{g_l^t}$ denotes the gradient of the l-th layer at time $t$, $\hat{g_l}$ denotes the EMA gradient, and $\gamma$ represents the momentum coefficient. For the sake of brevity, we shall exclude the superscript $t$ in the subsequent sections of this manuscript.
We take $\hat{g_l}$ as input and get the adaptation coefficient $\mu_l$:
\begin{equation}
\mu_l = \Omega(\hat{g_l}; \phi_l),
\end{equation}
where $\Omega(\cdot)$ is the chunking operation in \cite{ha2016hypernetworks} that partitions the gradient into uniformly-sized chunks. These segments are subsequently associated with the adaptation coefficients that are characterized by the trainable parameters $\phi_l$.
Specifically, the adaptation coefficient $\mu_l$ is composed of two components: a weight adaptation coefficient $\alpha_l$ and a feature adaptation coefficient $\beta_l$.
Then we conduct weight adaptation and feature adaptation.
The weight adaptation parameter $\alpha_l$ performs an element-wise multiplication on the corresponding weight of the backbone network, as described in:
\begin{equation}
\tilde{\theta_l} = tile(\alpha_l) \odot \theta_l,
\end{equation}
where we represent the feature maps of all channels in a TCN as $\theta_l$, while the adapted weights are denoted by $\tilde{\theta_l}$. The weight adaptor is applied per-channel on all filters using the tile function, which repeats a vector along the new axes, as indicated by $tile(\alpha_l)$. Finally, the element-wise multiplication is represented by $\odot$. Likewise, we have:
\begin{equation}
\tilde{h_l} = tile(\beta_l) \odot h_l,
\end{equation}
where $h_l = \tilde{\theta_l} \ast \tilde{h}_{l - 1}$ is the output feature map.

\subsection{Associative Memory}

In order to prevent a model from forgetting old patterns during continual learning in the context of time series, it is crucial to preserve the appropriate adaptation coefficients $\mu$, which encapsulate adequate temporal patterns for forecasting. These coefficients reflect the model's prior adaptation to a specific pattern, and thus, retaining and recalling the corresponding $\mu$ can facilitate learning when the pattern resurfaces in the future. Consequently, we incorporate an associative memory to store the adaptation coefficients of recurring events encountered during training. This associative memory is denoted as $M_l \in \mathbf{R}^{N\times d}$, where $d$ represents the dimensionality of $\mu_l$ and is set to a default value of 64.

\subsubsection{Memory Interaction Triggering. }
To circumvent the computational burden and noise that arises from storing and querying coefficients at each time step, FSNet propose to activate this interaction only when there is a significant change in the representation. The overlap between the current and past representations can be evaluated by taking the dot product of their respective gradients. FSNet leverage an additional EMA gradient $\hat{g'}_l$, with a smaller coefficient $\gamma'$ compared to the original EMA gradient $\hat{g}_l$, and measure the cosine similarity between them to determine when to trigger the memory. 
We use a hyper-parameter $\tau$, which we set to $0.7$, to ensure that the memory is only activated to recall significant pattern changes that are more likely to recur. The interaction is triggered when $cosine(\hat{g}_l, \hat{g'}_l) < - \tau$.

To guarantee that the present adaptation coefficients account for the entire event, which may extend over an extended period, memory read and write operations are carried out utilizing the adaptation coefficients of the EMA with coefficient $\gamma'$. The EMA of $\mu_l$ is computed following the same procedure as Equation~\ref{eq:ema}. In the event that a memory interaction is initiated, the adaptor retrieves the most comparable transformations from the past through an attention-read operation, which involves a weighted sum over the memory items:

\begin{equation}
\mathbf{r}_l = softmax(M_l\hat{\mu}_l),
\end{equation}
\begin{equation}
\tilde{\mu}_l = \sum_{i=1}^k TopK(\mathbf{r}_l)[i]M_l[i],
\end{equation}
where $TopK(\cdot)$ selects the top $k$ values from $\mathbf{r}_l$, and $[i]$ means the $i$-th element.
Retrieving the adaptation coefficient from memory enables the model to recall past experiences in adapting to the current pattern and improve its learning in the present. The retrieved coefficient is combined with the current parameters through a weighted sum: $\mu_l = \tau\mu_l + (1 - \tau)\tilde{\mu}_l$. Subsequently, the memory is updated using the updated adaptation coefficient:
\begin{equation}
M_l = \tau M_l + (1 - \tau)\tilde{\mu}\otimes TopK(\mathbf{r}_l),
\end{equation}
where $\otimes$ denotes the outer-product operator. So far, we can effectively incorporate new knowledge into the most pertinent locations, as identified by the top-k attention values of $\mathbf{r}_l$. Since the memory is updated by summation, it can be inferred that the memory $\mu_l$ does not increase as learning progresses.

\section{Datasets and Contextual Data}

\begin{figure}[htbp]
\centering
\includegraphics[width=.47\textwidth]{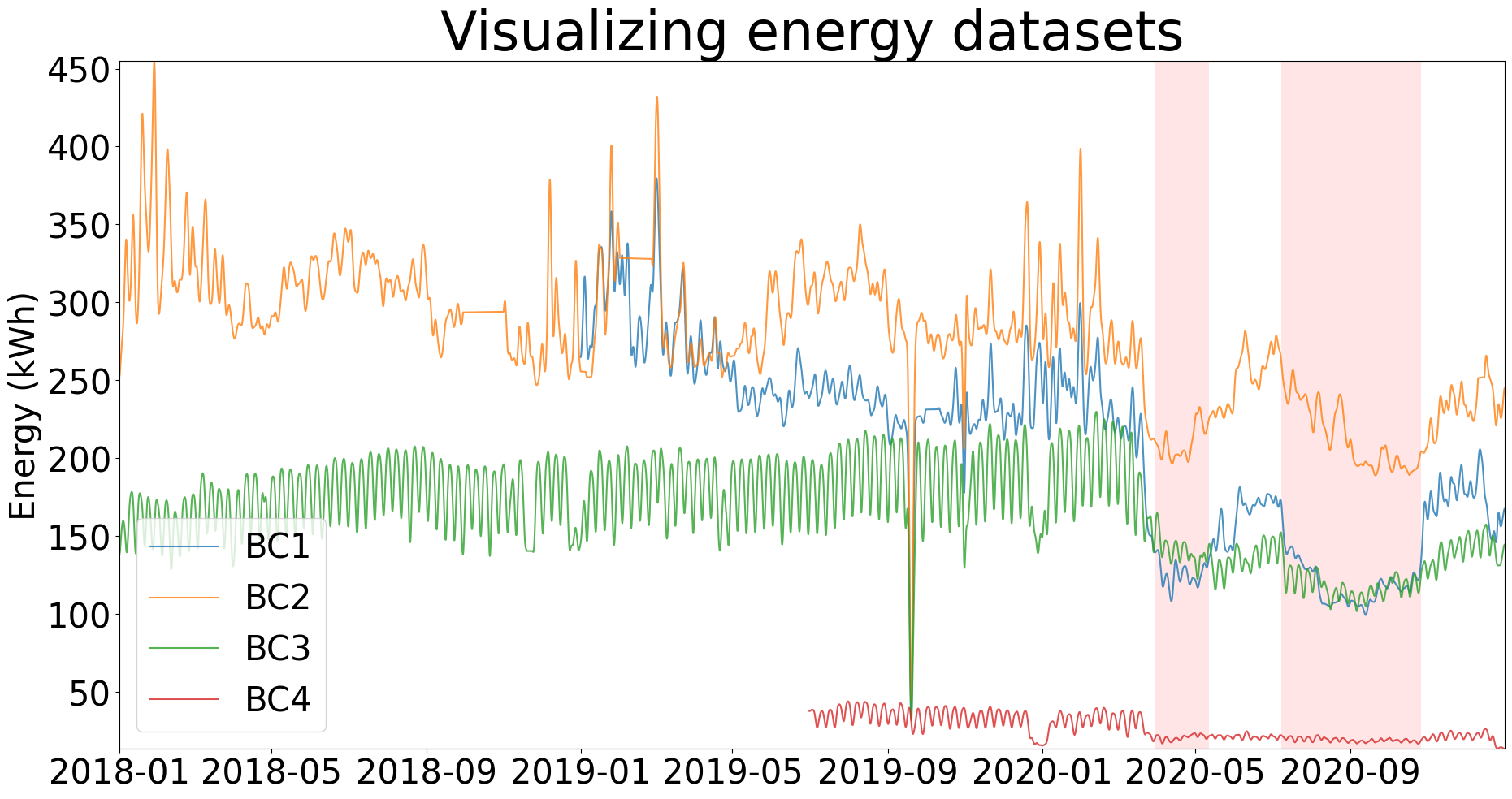}
\hfill
\includegraphics[width=.49\textwidth]{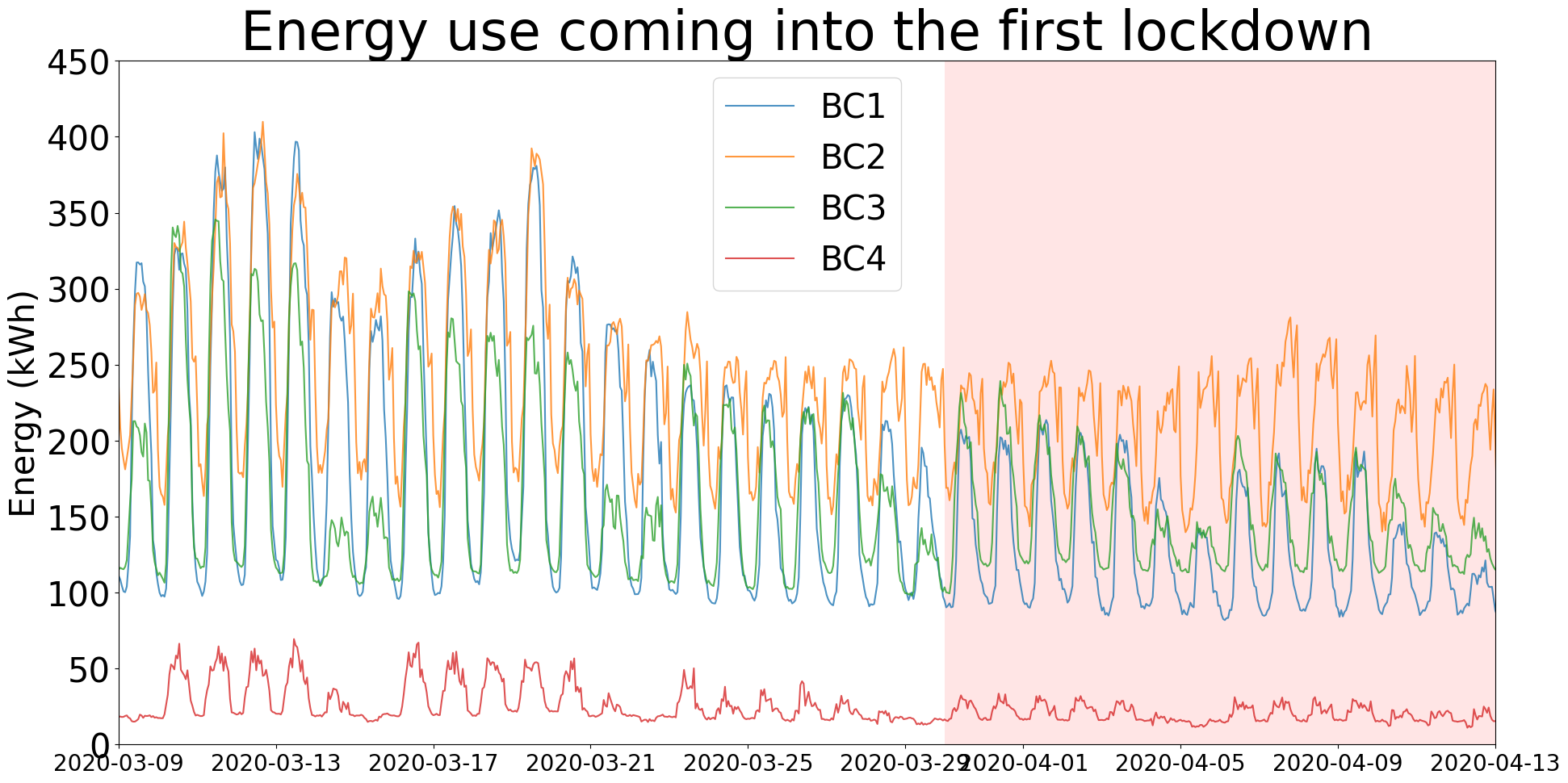}
\includegraphics[width=.47\textwidth]{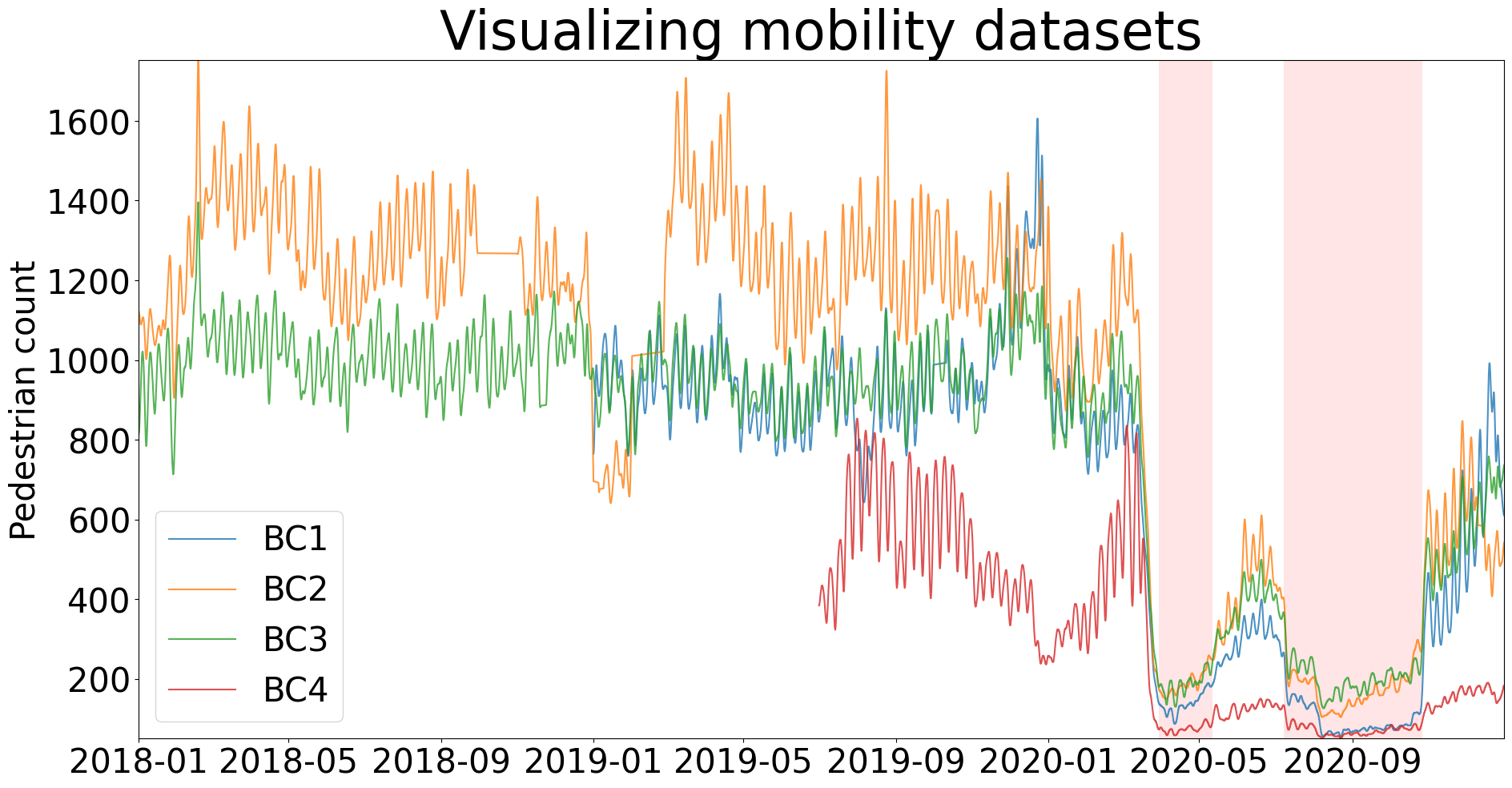}
\hfill
\includegraphics[width=.49\textwidth]{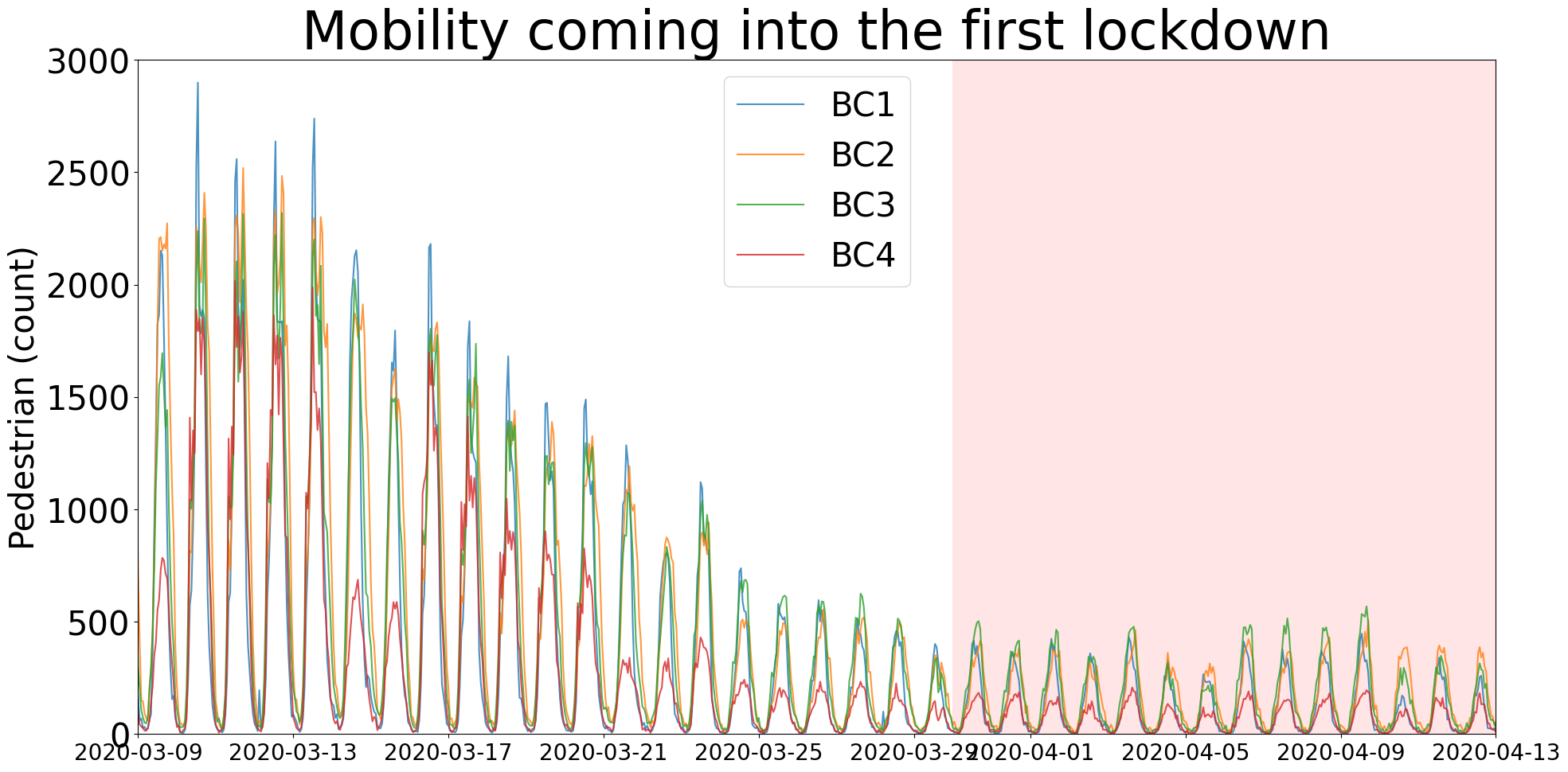}
\caption{
Visualizing the four datasets and their features, showing the significant changes in distributions due to lockdowns.
Plots on the left column are smoothed with a Gaussian filter with sigma = 24 hours.
Red areas are lockdowns.
}
\label{fig:ts_all_data}
\end{figure}

This paper is based on two primary data sources: energy usage data and mobility data, as well as two contextual datasets: COVID lockdown dates and temperature data. The statistical summary of the main datasets are provided in Table \ref{tab:data_stats} and visualized in Figure \ref{fig:ts_all_data}. These datasets were collected from four building complexes in the Melbourne CBD area of Australia between 2018 and 2021.

Table \ref{tab:data_stats} outlines the essential statistical properties of energy usage and mobility data collected from the four building complexes. It is evident from the data that energy usage varies significantly between the buildings, with BC2 having over ten times the average energy usage of BC4. Similarly, the mobility data shows distinct differences, with BC2 having a mean pedestrian count over three times greater than BC4. These differences emphasize the complexity of forecasting for energy usage in different building complexes.

\setlength{\tabcolsep}{6pt}
\begin{table}[htb]
\centering
\caption{The summary statistics of the four datasets, each of which represents an aggregated and anonymized building complex (BC).} \label{tab:data_stats}
\begin{tabular}{@{}cc|cccc@{}}
\toprule
\multicolumn{1}{l}{}                                                        &                  & \textbf{BC1} & \textbf{BC2} & \textbf{BC3} & \textbf{BC4} \\ \midrule
\multirow{5}{*}{\textbf{Temporal}}                                                   & start            & 2019-01-01   & 2018-01-01   & 2018-01-01   & 2019-07-01   \\
                                                                            & end              & 2020-12-31   & 2020-12-31   & 2020-12-31   & 2020-12-31   \\
                                                                            & num of record    & 17304        & 24614        & 26196        & 13200        \\
                                                                            & duration (years) & 2.0          & 3.0          & 3.0          & 1.5          \\
                                                                            & granularity      & hourly       & hourly       & hourly       & hourly       \\ \midrule
\multirow{7}{*}{\begin{tabular}[c]{@{}c@{}}\textbf{Energy}\\ (kWh)\end{tabular}}     & mean             & 207.27       & 278.17       & 166.42       & 26.64        \\
                                                                            & std              & 111.59       & 88.67        & 66.60        & 13.21        \\
                                                                            & min              & 3.25         & 1.32         & 5.38         & 0.00         \\
                                                                            & 0.25             & 112.72       & 203.38       & 112.62       & 17.45        \\
                                                                            & median           & 169.29       & 272.68       & 144.34       & 21.75        \\
                                                                            & 0.75             & 297.33       & 342.10       & 206.63       & 31.06        \\
                                                                            & max              & 611.67       & 709.41       & 371.64       & 83.01        \\ \midrule
\multirow{7}{*}{\begin{tabular}[c]{@{}c@{}}\textbf{Mobility}\\ (count)\end{tabular}} & mean             & 661.4        & 977.8        & 804.9        & 295.8        \\
                                                                            & std              & 876.8        & 936.2        & 761.4        & 387.3        \\
                                                                            & min              & 0            & 0            & 0            & 0            \\
                                                                            & 0.25             & 37           & 149          & 127          & 33           \\
                                                                            & median           & 209          & 614          & 528          & 135          \\
                                                                            & 0.75             & 1004         & 1818         & 1349         & 386          \\
                                                                            & max              & 6025         & 5053         & 3780         & 2984         \\ \bottomrule
\end{tabular}
\end{table}

It is worth noting that lockdown had a more significant impact on mobility than energy usage, as illustrated in Figure \ref{fig:ts_all_data}. Additionally, both energy usage and mobility started declining even before the start of lockdown.

\subsection{Energy Usage Data}

The energy usage data was collected from the energy suppliers for each building complex and measured the amount of electricity used by the buildings. To protect the privacy of the building owners, operators, and users, the energy usage data from each building was aggregated into complexes and anonymized. Buildings in the same complexes can have different primary use (e.g. residential, office, retails)

\subsection{Mobility Data}

\begin{figure}[!htb]
\centering
\includegraphics[width=.7\textwidth]{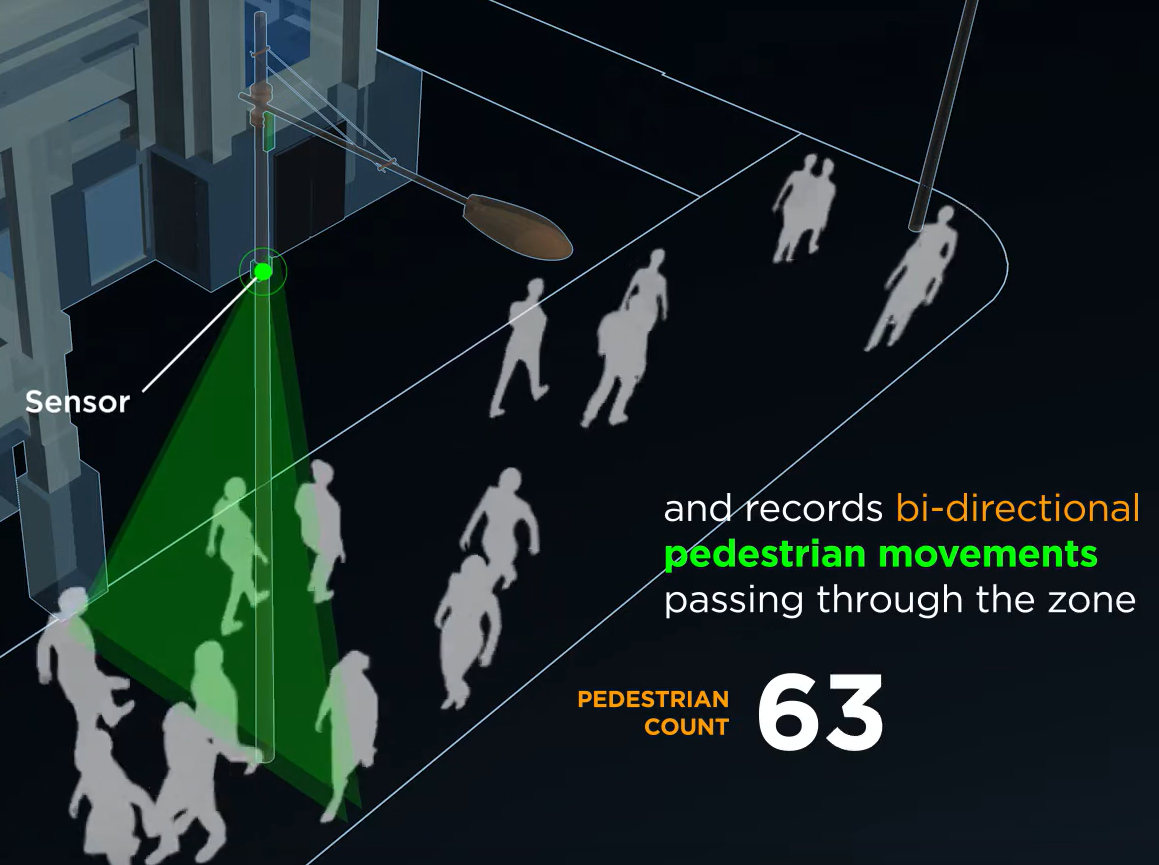}
\caption{
Diagram of automated pedestrian counting system. Obtained from the City of Melbourne website~\cite{CoM_peds}.
}
\label{fig:mob_sensor}
\end{figure}

The mobility data was captured by an automated pedestrian counting system installed by the City of Melbourne  \url{http://www.pedestrian.melbourne.vic.gov.au/}~\cite{CoM_peds}, and provided information on the movement patterns of individuals in and around each building complex. The system recorded the number of pedestrians passing through a given zone
as shown in Fig. \ref{fig:mob_sensor}. As no images were recorded, no individual information was collected. Some sensors were installed as early as 2009, while others were installed as late as 2021. Some devices were moved, removed, and upgraded at various times.
Seventy-nine sensors have been installed, and we have chosen four sensors, one for each building complex. We performed manual matching between the complexes and sensors by selecting the sensor that was closest to each building complex.

\subsection{COVID Lockdown Dates}

We used data on the dates of the COVID lockdowns in Melbourne, one of the strictest in the world. Our datasets coincides with the first lockdown from March 30, 2020 to May 12, 2020 (43 days), and the second lockdown from July 8 to October 27, 2020 (111 days). We also divided the time into pre-lockdown and post-lockdown periods, taking the date of the first lockdown (March 30, 2020) as the boundary. We took this information from \url{https://www.abc.net.au/news/2021-10-03/melbourne-longest-lockdown/100510710}~\cite{boaz_2021}.

\subsection{Temperature Data}
Temperature records are extracted from the National Renewable Energy Laboratory (NREL) Asia Pacific Himawari Solar Data~\cite{nrel_data}. 
As the building complexes are located in close proximity to one another, we utilized the same temperature data for all of them.

\subsection{Dataset Preprocessing}

For this study, we have fixed an observation of $L=24$ hours and a forecast horizon size of $H=24$ hours, to mimic a day-ahead forecasting experiment. 
To accurately link the foot traffic mobility data with the building, we carefully handpicked the pedestrian counting sensor that is located in the immediate vicinity of the building and used its corresponding mobility signal. The energy usage load of the building, the foot traffic volume, and the temperature degree were all aligned based on their timestamps.

\section{Experiments and Results}

We conducted two sets of experiments to evaluate the effectiveness of our proposed methods for predicting energy usage during anomalous periods.
The first set of experiments evaluated the impact of including mobility contextual data in our models.
The second set of experiments assessed the importance of continual learning.
In addition, we conducted ablation experiments on FSNet to investigate the impact of different components of the model on the overall performance.

\subsection{Experimental Setup}
The experiments were conducted on a high-performance computing (HPC) node cluster with an Intel(R) Xeon(R) Platinum 8268 CPU @ 2.90GHz and Tesla V100-SXM2. The software specifications included intel-mkl 2020.3.304, nvidia-cublas 11.11.3.6, cudnn 8.1.1-cuda11, fftw3 3.3.8, openmpi 4.1.0, magma 2.6.0, cuda 11.2.2, pytorch 1.9.0, python3 3.9.2, pandas 1.2.4, and numpy 1.20.0.

The data was split into three months for pre-training, three months for validation of the pre-training, and the rest was used for the usual continual learning setup. No hyperparameter tuning was conducted as default settings were used. The loss function used is MSE.

\subsection{Mobility}

\setlength{\tabcolsep}{5pt}
\begin{table}[htb]
\caption{
Performance comparison between different contextual features.
Results are average over 10 runs with different random seed.
The standard deviation is shown.
The algorithm used was FSNet with continual learning.
+M is the improvement of adding mobility over no context,
+T is the improvement of adding temperature over no context,
T+M is the improvement of adding mobility over temperature only.
}
\label{tab:res_mob}
\begin{tabular}{@{}cc|rrrrccc@{}}
\toprule
\multicolumn{1}{l}{\textbf{(MAE)}} & \textbf{dataset}     & \multicolumn{1}{c}{\textbf{\begin{tabular}[c]{@{}c@{}}no\\ context\end{tabular}}} & \multicolumn{1}{c}{\textbf{\begin{tabular}[c]{@{}c@{}}mobility\\ only\end{tabular}}} & \multicolumn{1}{c}{\textbf{\begin{tabular}[c]{@{}c@{}}temp.\\ only\end{tabular}}} & \multicolumn{1}{c}{\textbf{both}} & \textbf{+M}              & \textbf{+T}              & \textbf{T+M}             \\ \midrule
\multirow{8}{*}{\rotatebox[origin=c]{90}{Pre-Lockdown}}      & \multirow{2}{*}{BC1} & 0.1591                                                                            & 0.1587                                                                               & 0.1595                                                                            & 0.1516                            & \multirow{2}{*}{0.0004}  & \multirow{2}{*}{-0.0004} & \multirow{2}{*}{0.0079}  \\
                              &                      & $\pm$0.0252 & $\pm$0.0334 & $\pm$0.0269 & $\pm$0.0332 &  &  &  \\
                              & \multirow{2}{*}{BC2} & 0.1711 & 0.1993 & 0.1947 & 0.1708 & \multirow{2}{*}{-0.0282} & \multirow{2}{*}{-0.0236} & \multirow{2}{*}{0.0239} \\
                              &                      & $\pm$0.0085 & $\pm$0.0385 & $\pm$0.0391 & $\pm$0.0068 &  &  &  \\
                              & \multirow{2}{*}{BC3} & 0.2629 & 0.2866 & 0.2509 & 0.2403 & \multirow{2}{*}{-0.0237} & \multirow{2}{*}{0.0120} & \multirow{2}{*}{0.0105} \\
                              &                      & $\pm$0.0373 & $\pm$0.0534 & $\pm$0.0262 & $\pm$0.0095 &  &  &  \\
                              & \multirow{2}{*}{BC4} & 0.2706 & 0.2516 & 0.3142 & 0.2776 & \multirow{2}{*}{0.0190} & \multirow{2}{*}{-0.0436} & \multirow{2}{*}{0.0366} \\
                              &                      & $\pm$0.0370 & $\pm$0.0206 & $\pm$0.1581 & $\pm$0.0312 &  &  &  \\ \midrule
\multirow{8}{*}{\rotatebox[origin=c]{90}{Post-Lockdown}}     & \multirow{2}{*}{BC1} & 0.1484                                                                            & 0.1475                                                                               & 0.1434                                                                            & 0.1369                            & \multirow{2}{*}{0.0033}  & \multirow{2}{*}{0.0041}  & \multirow{2}{*}{0.0041}  \\
                               &                      & $\pm$0.0318 & $\pm$0.0464 & $\pm$0.0283 & $\pm$0.0355 &  &  &  \\
                               & \multirow{2}{*}{BC2} & 0.1636 & 0.1902 & 0.1849 & 0.1624 & \multirow{2}{*}{0.0072} & \multirow{2}{*}{-0.0194} & \multirow{2}{*}{0.0053} \\
                               &                      & $\pm$0.0085 & $\pm$0.0371 & $\pm$0.0381 & $\pm$0.0063 &  &  &  \\
                               & \multirow{2}{*}{BC3} & 0.2418 & 0.2654 & 0.2299 & 0.2198 & \multirow{2}{*}{-0.0014} & \multirow{2}{*}{-0.0251} & \multirow{2}{*}{0.0355} \\
                               &                      & $\pm$0.0374 & $\pm$0.0537 & $\pm$0.0252 & $\pm$0.0089 &  &  &  \\
                               & \multirow{2}{*}{BC4} & 0.3236 & 0.2943 & 0.4134 & 0.3282 & \multirow{2}{*}{0.0293} & \multirow{2}{*}{-0.1191} & \multirow{2}{*}{0.0852} \\
                               &                      & $\pm$0.0602 & $\pm$0.0294 & $\pm$0.3215 & $\pm$0.0502 &  &  &  \\ \bottomrule
\end{tabular}
\end{table}

To assess the significance of the mobility context in predicting energy usage during anomalous periods, we performed a contextual feature ablation analysis, comparing pre- and post-lockdown performance. Table \ref{tab:res_mob} presents the results of our experiments. Our findings suggest that the importance of mobility context is unclear in pre-lockdown periods, with mixed improvements observed, and the improvements are small compared to the standard deviations. However, post-lockdown, the importance of mobility context is more pronounced, and the best performance was achieved when both mobility and temperature contexts were utilized. Notably, our analysis revealed that post-lockdown, the improvement brought about by the mobility context is larger than that achieved through temperature alone, as observed in BC1, BC2, and BC4.
This could be due to the fact that temperature has a comparatively simple and regular periodic pattern such that deep learning models can deduce them from energy data alone.

\subsection{Continual Learning}

\begin{table}[htb]
\centering
\caption{
Comparing the performance of different algorithm with or without continual learning (CL).
The metric used is MAE.
Results are average over 10 runs with different random seed.
The standard deviation is shown.
}\label{tab:res_CL}
\begin{tabular}{@{}cc|rrrrrr@{}}
\toprule
&
\multirow{2}{*}{\textbf{dataset} } &
\multicolumn{1}{c}{\textbf{FSNet}} &
\multicolumn{1}{c}{\textbf{FSNet}} &
\multicolumn{1}{c}{\textbf{TCN}} &
\multicolumn{1}{c}{\textbf{OGD}} &
\multicolumn{1}{c}{\textbf{ER}} &
\multicolumn{1}{c}{\textbf{DR++}}
\\
& & \textbf{(no CL)} & & \textbf{(no CL)} & & &
\\ \midrule \midrule

\multirow{8}{*}{\rotatebox[origin=c]{90}{Pre-Lockdown}}
 & \multirow{2}{*}{BC1}
 	& 0.3703 & \textbf{0.1583} & 0.3668 & 0.2056 & 0.1820 & 0.1696 \\
 &  & $\pm$0.0607 & $\pm$0.0280 & $\pm$0.0379 & $\pm$0.0413 & $\pm$0.0217 & $\pm$0.0130 \\[1mm]
 & \multirow{2}{*}{BC2}
 	& 0.6272 & \textbf{0.1712} & 0.5176 & 0.2465 & 0.2322 & 0.2272 \\
 &  & $\pm$0.0914 & $\pm$0.0063 & $\pm$0.0607 & $\pm$0.0105 & $\pm$0.0056 & $\pm$0.0062 \\[1mm]
 & \multirow{2}{*}{BC3}
 	& 0.6750 & \textbf{0.2462} & 0.6500 & 0.3308 & 0.2862 & 0.2726 \\
 &  & $\pm$0.0638 & $\pm$0.0151 & $\pm$0.0698 & $\pm$0.0812 & $\pm$0.0432 & $\pm$0.0334 \\[1mm]
 & \multirow{2}{*}{BC4}
 	& 1.0018 & \textbf{0.2802} & 1.1236 & 0.3910 & 0.3511 & 0.3408 \\
 &  & $\pm$0.1053 & $\pm$0.0312 & $\pm$0.1040 & $\pm$0.0520 & $\pm$0.0323 & $\pm$0.0210 \\
 \midrule
\multirow{8}{*}{\rotatebox[origin=c]{90}{Post-Lockdown}}
 & \multirow{2}{*}{BC1}
 	& 0.4537 & \textbf{0.1429} & 0.4179 & 0.1797 & 0.1589 & 0.1482 \\
 &  & $\pm$0.0517 & $\pm$0.0275 & $\pm$0.0443 & $\pm$0.0342 & $\pm$0.0168 & $\pm$0.0094 \\[1mm]
 & \multirow{2}{*}{BC2}
 	& 0.6506 & \textbf{0.1628} & 0.5209 & 0.2313 & 0.2188 & 0.2148 \\
 &  & $\pm$0.0994 & $\pm$0.0057 & $\pm$0.0535 & $\pm$0.0085 & $\pm$0.0060 & $\pm$0.0068 \\[1mm]
 & \multirow{2}{*}{BC3}
 	& 0.7168 & \textbf{0.2255} & 0.7083 & 0.3014 & 0.2636 & 0.2518 \\
 &  & $\pm$0.0632 & $\pm$0.0145 & $\pm$0.0793 & $\pm$0.0709 & $\pm$0.0373 & $\pm$0.0286 \\[1mm]
 & \multirow{2}{*}{BC4}
 	& 1.8415 & \textbf{0.3314} & 1.8307 & 0.4496 & 0.4162 & 0.4043 \\
 &  & $\pm$0.2765 & $\pm$0.0520 & $\pm$0.2319 & $\pm$0.0643 & $\pm$0.0475 & $\pm$0.0338 \\
 \bottomrule
\end{tabular}
\end{table}

We conducted an experiment to determine the significance of continual learning by comparing the performance of various popular models with and without continual learning.

The models used in the experiment are:
\begin{itemize}
    \item \textbf{FSNet}~\cite{pham2022learning}: Fast and slow network, described in detail in the method section of this paper. In the `no CL', version we use the exact same architecture, however we use the traditional offline learning.
    \item \textbf{TCN}:~\cite{bai2018empirical}: Temporal Convolutional Network, is the offline learning baseline. It modifies the typical CNN using causal and dilated convolution which enhance its ability to capture temporal dependencies more effectively. The next three methods are different continual learning methods that uses TCN as the baseline.
    \item \textbf{OGD}: Ordinary gradient descent, a popular optimization algorithm used in machine learning. It updates the model parameters by taking small steps in the direction of the gradient of the loss function.
    \item \textbf{ER}~\cite{lin1992ER,chaudhry2019ER}: Experience Replay, a technique used to re-expose the model to past experiences in order to improve learning efficiency and reduce the effects of catastrophic forgetting.
    \item \textbf{DER++}~\cite{buzzega2020derpp}: Dark Experience Replay++ is an extension of the DER (Deep Experience Replay) algorithm, which uses a memory buffer to store past experiences and a deep neural network to learn from them. DER++ improves upon DER by using a dual-memory architecture, which allows it to store both short-term and long-term memories separately.
\end{itemize}
Table \ref{tab:res_CL} displays the results, which demonstrate the consistent importance of continual learning in both the pre- and post-lockdown periods, with improvements multiple times larger than the standard deviations.

\section{Conclusion}

In this study, we investigated the impact of mobility contextual data and continual learning on building energy usage forecasting during out-of-distribution periods. We used data from Melbourne, Australia, a city that experienced one of the strictest lockdowns during the COVID-19 pandemic, as a prime example of such periods. Our results indicated that energy usage and mobility patterns vary significantly across different building complexes, highlighting the complexity of energy usage forecasting. We also found that the mobility context had a greater impact than the temperature context in forecasting energy usage during lockdown. We evaluated the importance of continual learning by comparing the performance of several popular models with and without continual learning, including FSNet, OGD, ER, and DER++. The results consistently demonstrated that continual learning is important in both pre- and post-lockdown periods, with significant improvements in performance observed across all models. Our study emphasizes the importance of considering contextual data and implementing continual learning techniques for robust energy usage forecasting in buildings.

\subsubsection*{Acknowledgements}
We highly appreciate Centre for New Energy Technologies \href{https://c4net.com.au/}{(C4NET)} and Commonwealth Scientific and Industrial Research Organisation \href{www.csiro.au}{(CSIRO)} for their funding support and contributions during the project.
We would also like to acknowledge the support of Cisco's National Industry Innovation Network (NIIN) Research Chair Program.
This research was undertaken with the assistance of resources and services from the \href{https://nci.org.au/}{National Computational Infrastructure (NCI)}, which is supported by the Australian Government.
This endeavor would not have been possible without the contribution of Dr. Hansika Hewamalage and Dr. Mashud Rana.

\newpage

\section*{Ethics}

\subsection*{Ethical statement}

Data collection: the data used in this paper are a mixture of public and private data. For privacy reasons, the energy usage data cannot be made available publicly. The lockdown dates, and pedestrian data can be access publicly.
Lockdown dates is by ABC, an Australian public news service \url{https://www.abc.net.au/news/2021-10-03/melbourne-longest-lockdown/100510710} and the pedestrian data is from City of Melbourne, a municipal government \url{http://www.pedestrian.melbourne.vic.gov.au/}.

Statement of Informed Consent: This paper does not contain any studies with human or animal participants. There are no human participants in this paper, and informed consent is not applicable.

\subsection*{Ethical considerations}

There are several ethical considerations related to this paper.

\subsubsection*{Data privacy}
The use of data from buildings may raise concerns about privacy, particularly if personal data such as occupancy patterns is being collected and analyzed. Although the privacy of individual residents, occupants, and users  are protected through the building level aggregations, sensitive information belonging to building managers, operator, and owners might be at risk. To this end, we choose to further aggregate the few buildings into complexes and make it anonymous. Unfortunately, the implication is that we cannot publish the dataset.

\subsubsection*{Bias and discrimination}
There is a risk that the models used to predict energy usage may be biased against certain groups of people, particularly if the models are trained on data that is not representative of the population as a whole. This could lead to discriminatory outcomes, such as higher energy bills or reduced access to energy for marginalized communities. We do acknowledge that the CBD of Melbourne, Australia is not a representative of energy usage in buildings in general, in CBD around the world, nor Australia. However, our contribution specifically tackle the shift in distributions, albeit only temporally and not spatially. We hope that our contributions will advance the forecasting techniques, even when the distributions in the dataset are not representative.
\subsubsection*{Environmental impact}
This paper can make buildings more sustainable by improving energy usage forecasting, even during anomalous periods, such as the COVID-19 pandemic. Robust and accurate forecasting enables building managers to optimize energy consumption and reduce costs. By using contextual data, such as human mobility patterns, and continual learning techniques, building energy usage can be predicted more accurately and efficiently, leading to better energy management and reduced waste. This, in turn, can contribute to the overall sustainability of buildings and reduce their impact on the environment.

\newpage

%
%
%
\bibliographystyle{splncs04}
\bibliography{1bib}

\end{document}